\documentclass[sigconf]{acmart}

\copyrightyear{2020}
\acmYear{2020}
\setcopyright{acmcopyright}
\acmConference[HRI '20]{Proceedings of the 2020 ACM/IEEE International Conference on Human-Robot Interaction}{March 23--26, 2020}{Cambridge, United Kingdom}
\acmBooktitle{Proceedings of the 2020 ACM/IEEE International Conference on Human-Robot Interaction (HRI '20), March 23--26, 2020, Cambridge, United Kingdom}
\acmPrice{15.00}
\acmDOI{10.1145/3319502.3374812}
\acmISBN{978-1-4503-6746-2/20/03}

\usepackage{siunitx}
\usepackage{subcaption}
\usepackage{cleveref}
\usepackage[utf8]{inputenc}  

\usepackage{algorithm2e}

\begin{document}
\fancyhead{}
 
\title{Would You Mind Me if I Pass by You?}
\subtitle{Socially-Appropriate Behaviour for an Omni-based Social Robot in Narrow Environment}


\author{Emmanuel Senft}
\affiliation{%
  \institution{ATR}
  \city{Kyoto}
  \country{Japan}
  }
\affiliation{%
  \institution{University of Wisconsin-Madison}
  \city{Madison}
  \state{WI}
  \country{USA}
  }
\email{esenft@wisc.edu}
\author{Satoru Satake}
\affiliation{%
  \institution{ATR}
  \city{Kyoto}
  \country{Japan}
  }
\email{satoru@atr.jp}
\author{Takayuki Kanda}
\affiliation{%
  \institution{ATR}
  \city{Kyoto}
  \country{Japan}
  }
\affiliation{%
  \institution{Kyoto University, Japan}
  \city{Kyoto}
  \country{Japan}
  }
\email{kanda@i.kyoto-u.ac.jp}

%\renewcommand{\shortauthors}{Senft, et al.}

\begin{abstract}
  Interacting physically with robots and sharing environment with them leads to 
  situations where humans and robots have to cross each other in narrow corridors.
  In these cases, the robot has to make space for the human to pass. 
  From observation of human-human crossing behaviours, we isolated two main factors in
  this avoiding behaviour: body rotation and sliding motion. We implemented a robot
  controller able to vary these factors and explored how this variation impacted on people's perception. Results from a
  within-participants study involving 23 participants show that people prefer 
  a robot rotating its body when crossing them. Additionally, a sliding motion
  is rated as being warmer. These results show the importance of social avoidance
  when interacting with humans. 
\end{abstract}



%, human modelling}

\maketitle

\section{Introduction}

In the last few years, robots have started to enter into human spaces. Today, social robots are being mass-produced and deployed in hotels \cite{ivanov2017adoption}, hospitals \cite{mutlu2008robots}, or other places like stores and shopping malls \cite{K5}. Robots now navigate in human environments.

Previous research studied ways for robots to navigate in human space (e.g. \cite{kruse2013human,rios2015proxemics}). Although robots are capable to produce collision free paths, they will not be socially accepted if people feel uncomfortable about them.
Hence, many recent HRI studies investigate socially-acceptable navigation, exploring for instance how a robot should approach humans to avoid scaring them \cite{sisbot2007human}, or avoiding them early to maintain social distance \cite{shiomi2014towards}. 

\begin{figure}
    \centering
		\includegraphics[width=0.4\columnwidth]{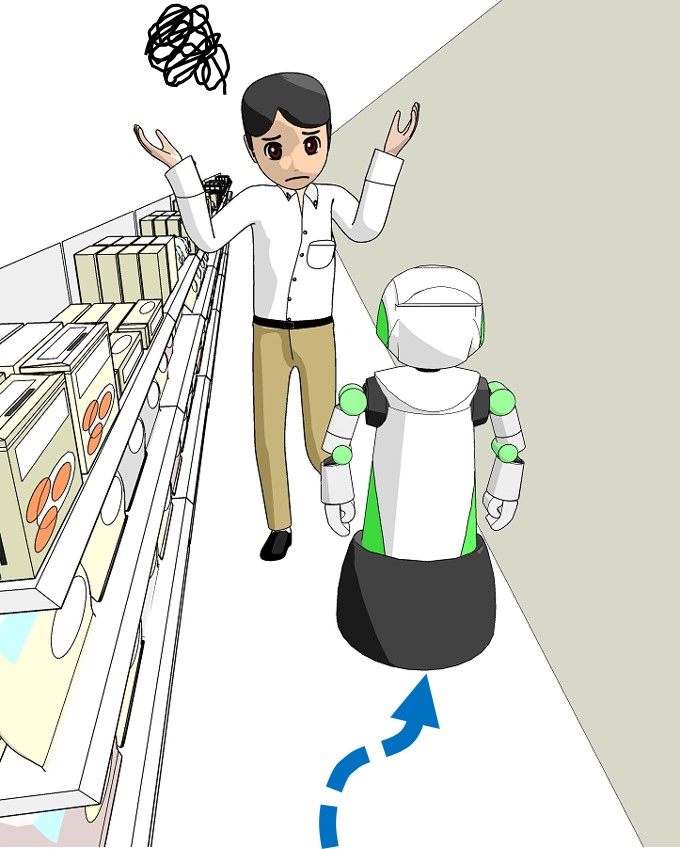}
    \caption{Robot in a narrow corridor}
    \label{fig:intro}
\end{figure}

However, unlike wide open spaces, some in-door environments such as hotels and stores have long narrow corridors where crossing can become an issue. For instance, typical stores include many of such narrow corridors, and a robot serving as a shop-clerk would need to operate carefully.
In this situation, when a robot is moving and a customer desires to cross the same corridor (cf. Fig. \ref{fig:intro}), the ideal robot's behaviour is still unclear.

As illustrated in the figure, the robot could move aside, open the space for the customer, and wait for the customer to pass by. Perhaps, this is the best behaviour produced by a typical collision-avoiding planner for a classic robot, i.e. a robot with two-degrees of freedom on its mobile base like a differential-drive motors (e.g. Pioneer 3-DX and Create \cite{Create} ).

In contrast, we human beings have more degrees of freedom in our movements and we use them to behave socially. As we will discuss in Sections \ref{sec:background} and \ref{sec:pilot}, people change their body orientation to yield the way in a nicer manner. With such extra degrees of freedom in a locomotion base, could a robot have a more socially-acceptable behaviour? Commercially-available robots, like Pepper \cite{Pepper}, start to have omni-directional bases, which add an extra degree of freedom in the motion. Thus like a person, such robots could change their body orientations while they navigate. More complex locomotion systems, such as bipedal robots, could also replicate this effect, but as of today, no such system is commercially available.

In this research, we first investigated how people typically avoid each other in a narrow corridor space. From the analysis, we identified the factors that could contribute to the socially-acceptable and efficient passing behaviour. That is, {\it body rotation} and {\it sliding}. We then implemented a path-planner for an omni-directional robot that timely realises these factors in a human-like way. Finally, we conducted a user study to investigate the effect of these factors.

\section{Background}
\label{sec:background}
\subsection{Human Behaviour in Crossing Interaction}
\label{sec:background-humanBehaviour}
Scientists have explored spatial behaviour of humans in crowded space for multiple decades now. The study of proxemics, ``the interrelated observations and theories of humans use of space as a specialized elaboration of culture'' \cite{hall1910hidden} is a vibrant topic of research in psychology and sociology. Moreover, basic concepts such as personal space \cite{CaplanPersonalSpace} and formation \cite{Kendon} have been investigated.

Some research especially investigate pedestrian crossing behaviours. Typically, previous studies had a major focus on people's trajectory, and addressed how people avoid each other. Helbing and his colleagues' seminal work modelled people's behaviour based on virtual social `forces', as the way people avoid resembles to the situation where physical repulsive forces are applied to nearby others \cite{HelbingSFM}. While initial studies had high density crowds, like panic and escapes situations, recent studies also covers various other situations such as a wide corridor in a shopping mall \cite{zanlungo2011social}. Overall, these studies mostly focus on people trajectories.

In contrast, there is a relatively small number of studies that addressed the problem of crossing behaviours in narrow environments. In such a situation, more precise behaviours, such as how people control their upper torso, can be as important as people's trajectories. 
In \cite{collett1974patterns}, Collett et al. observed humans moving on a pedestrian crossing and analysed their behaviours when taking part in \emph{passes}, the action of passing by a partner. Collett et al. report findings similar to \cite{wolf1973notes}: people tend to use a \emph{step-and-slide} strategy to avoid people when crossing them. Wolf defines this step-and-slide as ``a slight angling of the shoulders and an almost imperceptible side step''. This strategy mostly happens when crossing in narrow environments, when people's trajectories would overlap. Both parties rotate their shoulders to make enough space to cross \cite{yamamoto2019body}. 

\subsection{Related Studies in Human-Robot Interaction}
Motion planning and obstacle avoidance are well established techniques in robotics. For instance, Dynamic Window Approach (DWA) \cite{fox1997dynamic} is one of the commonly-used techniques for robotic planning that can deal with dynamic obstacles. Using such a technique, we could generate a collision-free path if we treat people as dynamically moving obstacles.

However, people are not dynamic obstacles. Even if a robot is able to safely avoid people without any collisions, people would not socially accept robots if the robots' motions make them feel uncomfortable. Hence, recent advances in human-robot interaction (HRI) started to consider the social side of navigation in human environments \cite{kruse2013human,mead2017autonomous,rios2015proxemics}. 
For instance, one of the pioneering work was conducted by Sisbot et al. They first modelled situations where people feel uncomfortable with the robot's navigation (e.g. a robot suddenly appearing from an area hidden by obstacles). Subsequently, they proposed a planner avoiding paths that could lead to such uncomfortable situations \cite{sisbot2007human}.

Socially appropriate behaviours of robots were also studied in the context of passing people.
The basic approach is to find a collision-free path allowing the robot to reach its goal without colliding with the human.
Researchers have augmented this basic approach with information such as the velocity of the incoming pedestrian \cite{spalanzani2012risk}, taking into account some social norms \cite{kirby2009companion}, modelling joint strategies followed by groups of people \cite{kuderer2012feature}, using social forces \cite{ferrer2013robot}, using qualitative trajectory calculus to formalise the crossing problem \cite{hanheide2012analysis}, or handling dynamic environments \cite{pandey2010framework}.

However, these approaches generally focus on finding a path and assume the robot's orientation to follow the direction of the path, as it is forced for differential robots. However, with the advances in robot locomotion, and especially the democratisation of omni-directional platforms, robots have the opportunity to uncouple their body orientation and their moving direction. This opens new potential of behaviours seldom explored by the community.

\section{Observing Human-Human Crossing Interaction} \label{sec:pilot}
To find people's common strategy when they cross in a narrow corridor, we conducted an observation of human-human crossing interaction.

\subsection{Environment and Procedure}
We conducted an observatory data-collection in a simulated store at our laboratory. We asked pairs of other laboratory members, who did not know our research purpose, to walk in this simulated store. We instructed them to role-play a scenario in a store, where one was acting as a shop clerk and the other was a customer, and asked them to repeatedly cross each other in narrow corridors. With these procedure, we collected the data of 54 pairs of crossing behaviours. 

\subsection{Observation Results}
We examined the observed crossing behaviours, and classified them into typically-observed patterns. We found three crossing patterns: {\it step - slide - rotate}, {\it step - no slide - rotate}, and {\it step - no slide - no rotate}. Table \ref{tab:obs} shows the number of times each pattern was observed in our study. We report each of pattern in subsequent sections.

\begin{table}[htb]
	\caption{Observed behaviours in human-human crossing}
	\begin{tabular}{lrr|r}
		\hline
		Category & Shop clerk & Customer & Total \\
		\hline
{\it step - slide - rotate} & 53 & 46 & 99\\
{\it step - no slide - rotate} & 1 & 2 & 3\\
{\it step - no slide - no rotate} & 0 & 6 & 6\\
		\hline
	\end{tabular} 
	\label{tab:obs}
\end{table}

\subsubsection{Step - slide - rotate}
\label{sec:step-slide-rotate}
\begin{figure}
    \centering
	\begin{subfigure}[b]{0.15\textwidth}
	    \centering
		\includegraphics[width=.9\textwidth]{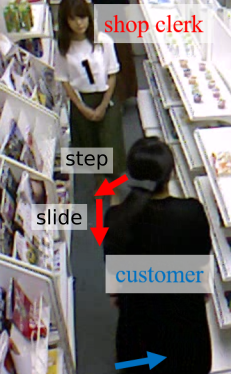}
		\caption{Both {\it step} aside and start to {\it slide}.}
    \label{fig:InitRota}
	\end{subfigure}
	\begin{subfigure}[b]{0.15\textwidth}
	    \centering
		\includegraphics[width=0.9\textwidth]{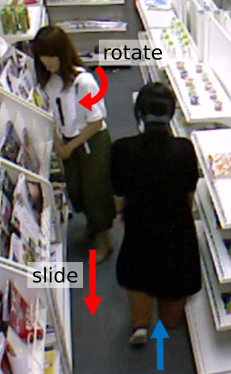}
		\caption{Just in time, shop clerk {\it rotates}.}
    \label{fig:InitRotb}
	\end{subfigure}
	\begin{subfigure}[b]{0.15\textwidth}
	    \centering
		\includegraphics[width=0.9\textwidth]{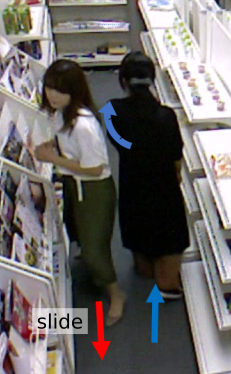}
		\caption{Both continue to {\it slide}.}
    \label{fig:InitRotc}
	\end{subfigure}
	\caption{An example of {\it step-slide-rotate} behaviour.}
    \label{fig:InitRot}
 \end{figure}

The {\it step - slide - rotate} sequence (Fig. \ref{fig:InitRot}) is the most frequently occurring pattern.
Firstly, a person takes a {\it step} to move aside and yield space for the oncoming person (Fig. \ref{fig:InitRota}).
The {\it step} phase was sometimes omitted when both people already walked on opposite sides of the corridor.
After the {\it step}, she started to {\it slide}, i.e. she continued to move forward unless the oncoming person came too close and created a risk of collision.
For instance, if the coming person walked on the same side, the person stopped to wait for the coming person to change her course, or had a {\it step} on the other side again to yield the way.  
Finally, she {\it rotated} her body (Fig. \ref{fig:InitRotb}) just in time before the moment of crossing and then continued the slide.

The {\it rotate} sequence was commonly started by the person who should prioritise the other one (i.e. the one with the shop clerk role).
In Fig. \ref{fig:InitRotb}, the shop clerk rotated her body first, largely before the other one started the body rotation. 
Then, the customer slightly rotated her body and they finally passed by each other (Fig. \ref{fig:InitRotc}).

\subsubsection{Step - no slide - rotate}
\begin{figure}
    \centering
	\begin{subfigure}[b]{0.2\textwidth}
	    \centering
		\includegraphics[width=0.72\textwidth]{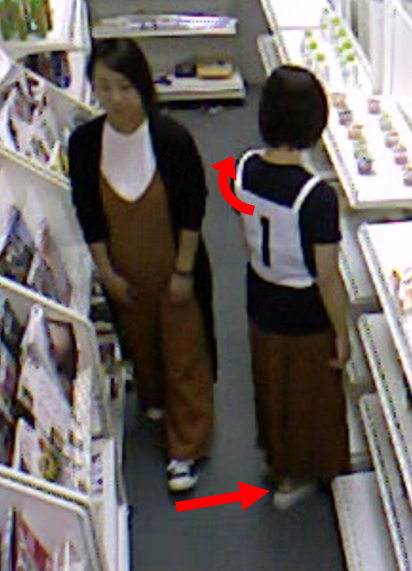}
		\caption{{\it step - no slide - rotate}.}
    \label{fig:otherCrossinga}
	\end{subfigure}
	\begin{subfigure}[b]{0.2\textwidth}
	    \centering
		\includegraphics[width=0.64\textwidth]{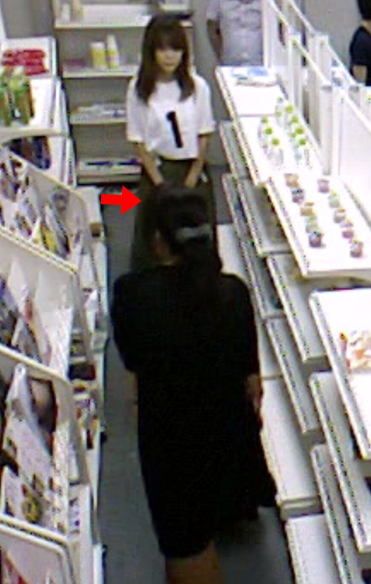}
		\caption{{\it step - no slide - no rotate}.}
    \label{fig:otherCrossingb}
	\end{subfigure}
    \caption{Other crossing behaviours.}
    \label{fig:otherCrossing}
 \end{figure}

Fig. \ref{fig:otherCrossinga} shows an example of {\it step - no slide - rotate} sequence. During crossing, one of the person stopped to wait the oncoming person to pass without {\it slide} motion.

This pattern was rarely observed; it only happened when both persons kept walking on the same side after their {\it step} behaviour; in that case, one stopped and waited for the other to pass by. 

\subsubsection{Step - no slide - no rotate}
This pattern was also rarely observed. In this pattern, during the crossing, one of the person waited without sliding motion, and did not rotate her body while crossing (Fig. \ref{fig:otherCrossingb}).
This {\it step - no slide - no rotate} sequence was only observed for customers when on the same side as the clerk. In this case, the customer might stop and wait for the clerk to yield the way.

\subsection{Modelling of Step - Slide - Rotate Behaviour}
\begin{figure}
    \centering
    \includegraphics[width=0.3\textwidth]{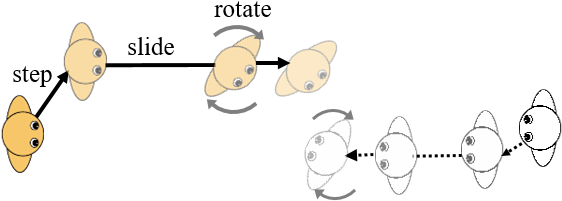}
    \caption{The model of {\it step - slide - rotate} sequence.}
    \label{fig:model}
 \end{figure}

From our observations, we found that the {\it step - slide - rotate} behaviour is the main one used for crossing in a narrow space, hence we developed a model for this behaviour (summarised in Fig. \ref{fig:model}). 
This model consists of three sequences: {\it step}, {\it slide}, and {\it rotate}.
In the {\it step} phase, the person moves to the side of the corridor to free the way to the oncoming person.
In the {\it slide} phase, the person keeps moving forward until the distance to the oncoming person is close enough to pass by.
In the {\it rotate} phase, the person rotates her body just in time, %before passing through
while continuing to slide.
The person prioritising the oncoming person starts the {\it rotate} phase before the other one does.

\section{System and Implementation}
 \label{sec:system}
We implemented an autonomous robot system that reproduces the step - slide - rotate  behaviour observed in Section \ref{sec:pilot}.

\subsection{Robot Hardware}
We used the upper body of Robovie-R3, which is characterised by its human-like physical appearance with a height of 110 cm and a width of 54 cm (cf. Fig. \ref{fig:photo}). We used QFS-02-ver1 developed by Qfeeltech for its mobile base. It has omni-directional wheels that allow it to move in any direction at a maximum speed of 1.5 m/sec. Its maximum angular velocity is 1 rad/sec (around 57$^\circ$/sec). The robot is equipped with a LiDAR (Velodyne HDL-32E) at a height of 143 cm, which is used for localisation and people-tracking. It also has four laser-range-finders (HOKUYO UTM-30LX) on its bottom, which are used for collision avoidance.

\subsection{Architecture}
\begin{figure}
    \centering
    \includegraphics[width=.95\columnwidth]{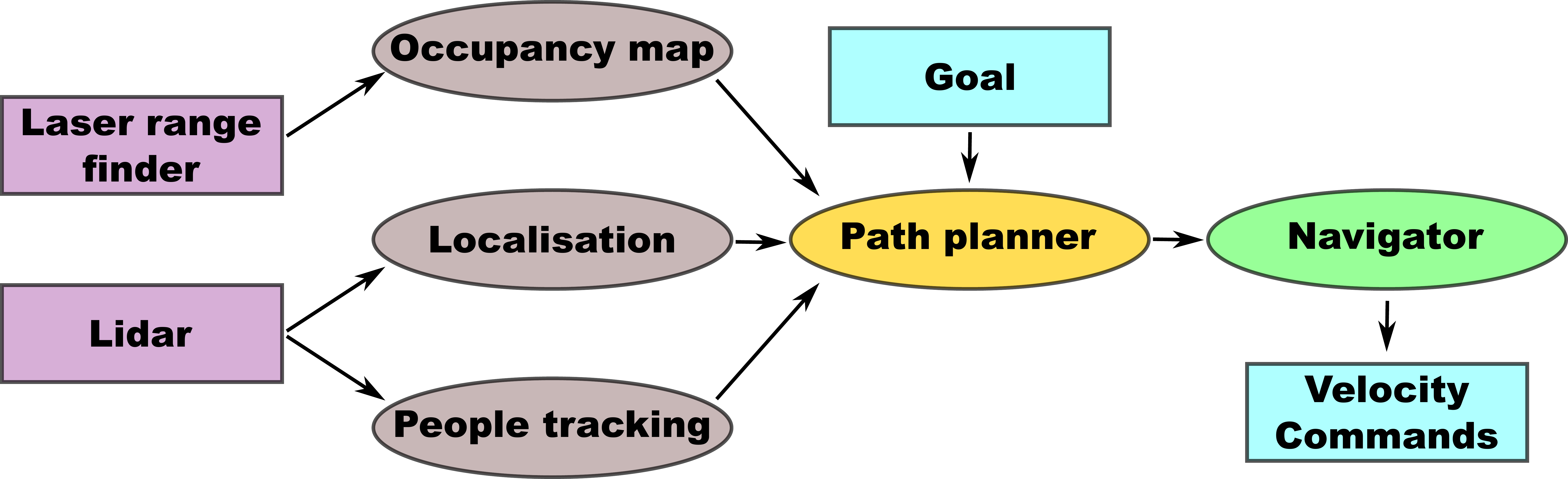}
    \caption{Information flow to generate velocity commands.}
    \label{fig:arch}
\end{figure}

Fig. \ref{fig:arch} shows the inputs and outputs of the path planner, the main module in our control architecture. Three modules (occupancy map, localisation and people tracking) are responsible to gather inputs from the sensors (laser-range-finder and LiDAR), process them and feed them to the planner.
When a goal position is provided, the path planner generates a path to reach that goal.
If there is no people on the way, the planner generates a straight line. However, if there is a person, the planner will create a path to avoid them, with a step, a slide and a rotating motion.
This path includes both position and orientation for the robot, and is updated at 10Hz.

\subsection{Step-Slide-Rotate Planner}
\label{sec:plan}

Our path planner can run in two modes: free-navigation and people-avoidance.
When no one is between the robot and its goal, the planner is in the free-navigation mode, and generates a simple path composed of two points: the current position and the current goal.
When a person is detected between the robot and its goal, the robot enters the people-avoidance mode. 
This mode generates a motion composed of two phases: the step phase and the slide-rotate phase.

\subsubsection{Step phase}
The first phase of this motion is the {\it step} phase, when the robot moves towards a wall to free space in the corridor for the oncoming human. The timing to start this step is important: 
a late step could be missed by the people passing by, or could not allow the robot to complete its body rotation later on, and 
an early step could lose context, preventing people to understand the robot's intention. In early tuning phases with other laboratory members (naive about the study goal), we tested multiple values for parameters. We observed participants' reaction and conducted small interviews to tune our parameters. Based on these observations, we empirically decided to start the step when people were less than four meters away. This value was robust to typical variations for velocity, acceleration and trajectories.

The side of the step is decided by the position of the incoming person.
If the person is on the left, the robot would go to the right and vice-versa. The first part of the step is slowing down, and then the robot moves in diagonal as shown in Fig. \ref{fig:model}.
In this phase, the omni-directional mobile base of the robot allows it to move on the side while facing the corridor, which produces a more natural behaviour compared to a classic robot with a differential-drive base.

\subsubsection{Slide-rotate phase}
After the {\it step} phase, the planner enters in the {\it slide-rotate} phase. In this phase, the planner generates a wall following trajectory (i.e. {\it slide} motion) together with a {\it rotate} motion of its body when people are about to cross the robot. Controlling the orientation independently of the trajectory is the key for this phase and is only possible due to the omni-directional mobile base.

\paragraph{Slide motion:}
During the slide-rotate phase, the planner generates a temporary goal along the wall, on the same side as the \emph{step} and half a meter in front of the robot. This moving target allows the robot to follow the wall closely in a smooth manner. During this motion, the robot is consistently facing forward.

\paragraph{Rotate motion:}
One key part of the slide-rotate phase is the rotation. The robot has to rotate its body just on time before the crossing. As robots are usually expected to prioritise people, we need this rotation to happen before the oncoming person has to rotate their body (based on our observation in Sec. \ref{sec:step-slide-rotate}).%}

\begin{figure}
    \centering
	\begin{subfigure}[b]{0.15\textwidth}
	    \centering
		\includegraphics[width=0.6\textwidth]{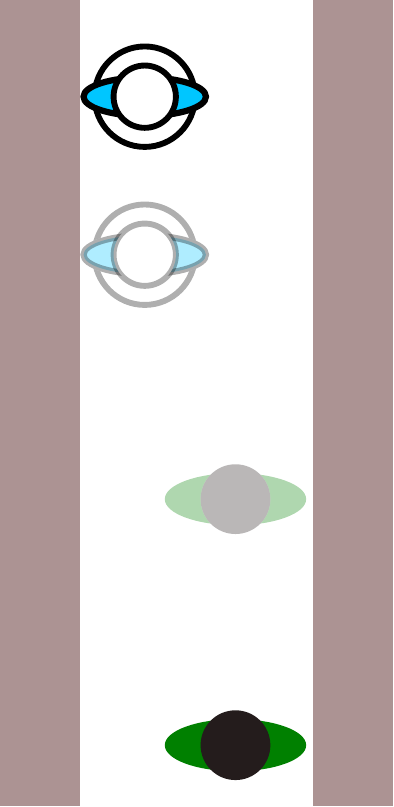}
		\caption{Robot and human still far apart.}
	\end{subfigure}
	\begin{subfigure}[b]{0.15\textwidth}
	    \centering
		\includegraphics[width=0.6\textwidth]{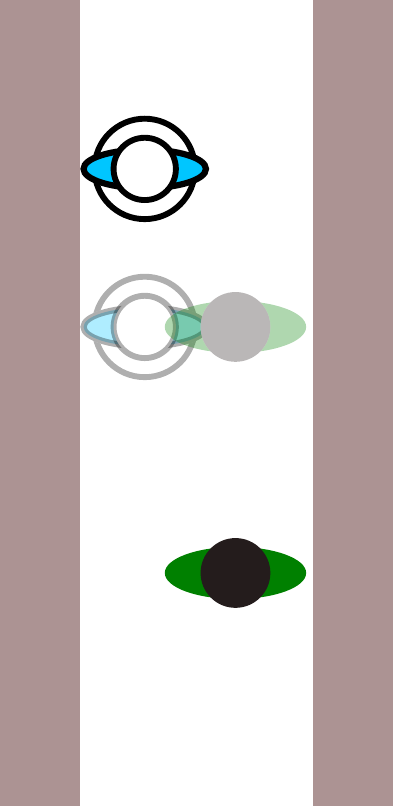}
		\caption{1.8s before crossing, starts rotation.}
	\end{subfigure}
	\begin{subfigure}[b]{0.15\textwidth}
	    \centering
		\includegraphics[width=0.6\textwidth]{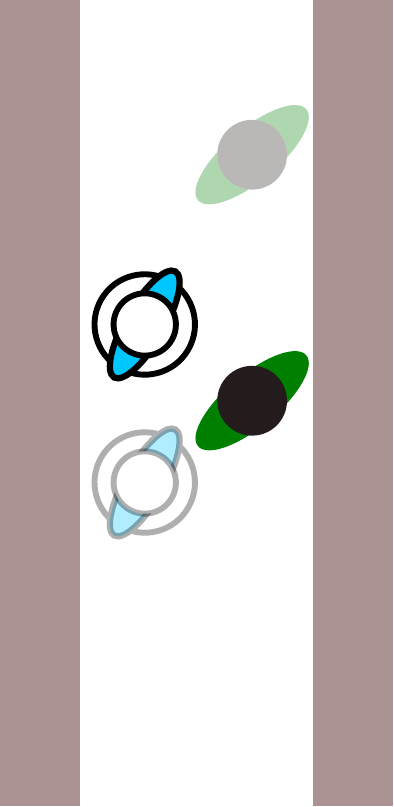}
		\caption{Crossing near, rotation finished.}
	\end{subfigure}
    \caption{Diagrams of the crossing interaction. Faded images represent the prediction of the robot and the human position after 1.8 seconds. 
    }
    \label{fig:pred}
\end{figure}

To achieve this \emph{just in time} rotation, the robot estimates the time before crossing by using the decrease of the relative distance between the human and the robot (Fig. \ref{fig:pred}): 
$t_{cross} = d_p / v_r$,
where $d_p$ indicates the distance between the robot the oncoming person, and $v_r$ the relative speed between the robot and the person. 

The planner then compares this crossing time to a threshold (empirically set during the tuning phase to 1.8s).
If the crossing time is inferior, the rotation motion is triggered, turning the robot's body of 60$^\circ$.
This timing allowed our system to start the rotation early enough to give people the opportunity to see the rotation, and give the robot enough time to complete the rotation right before the crossing actually happens.

We decided that the robot would perform an ``open'' rotation (i.e. turning its body towards the centre of the corridor and the incoming person) as participants accepted it well (they liked to see the robot's face) in our empirical trials.

\subsection{Other modules}

\subsubsection{Navigator}
The navigator's task is to reach the first goal of the path provided by the planner. To do so, the navigator generates a set of velocity commands (x, y and theta) using a modified PID coupling the directions x and y to ensure straight lines, using the integration only when approaching goals, taking into account the maximal speed and acceleration of the robot and avoiding direct collisions with obstacles. % and keeping centre of the corridor if not in a people avoiding motion.

\subsubsection{Localisation}
Even in small environment, robots need to know precisely their position to ensure a correct trajectory. This localisation is often not possible with odometry alone. Consequently, to
obtain the robot's position, we applied a particle-filter-based method on the LiDAR and odometry inputs. 
For each particle, we conducted map-matching using an end point model for computing the concordance. The module periodically corrects the robot's location at 10Hz. The localisation module uses a 3-dimensional map prepared in advance. This map was created by teleoperating the robot in the environment and applying 3D Toolkit SLAM library to match consecutive scans (containing lidar 3D scans, odometry data and IMU information) and performing global relaxation \cite{borrmann2008globally}.

\subsubsection{People-tracking}
We implemented a people-tracking algorithm for the point cloud from LiDAR. It consists of three process steps: background subtraction, people detection, and people-tracking. Background subtraction is done immediately after the localisation, compares the map and the raw point cloud, and removes the entities recorded in the map. After this, the remaining point cloud shows the movable entities. The people detection step, applies clustering and detects entities that are human-size. Finally, the people-tracking step applies a particle filter to the detection result, so that even with occlusions, e.g. people passing by, it can continuously track the people's locations. In our setting, the system tracked people who were within 20 m of the robot at 10Hz.

\section{Experiment} \label{sec:method}
We designed an experiment to evaluate the impact of the robot's body rotation and sliding motion on participants when crossing with it. We used a shopping scenario where participants had to pick up grocery items in a shop inhabited by a robot shop keeper. 

\subsection{Hypotheses and Prediction}
As described in the Section \ref{sec:system}, the human \emph{step - slide - rotate} behaviour is composed of two factors: the rotation and the sliding motion. We designed hypotheses to explore the impact of these two factors.

\subsubsection{Hypotheses on rotation}

As show in our observation (Sec. \ref{sec:pilot}) and previous work \cite{collett1974patterns}, in narrow spaces, humans often use body rotation to make more space to people coming towards them. 
Consequently, we posit that by reusing human social behaviours, a robot rotating its body when crossing would be perceived more positively (e.g. more caring, social, and kind) 
than a robot not doing this. In social interactions, these factors: care, sociability, kindness can be grouped under the \emph{warmth} label \cite{cuddy2011dynamics,carpinella2017robotic}. Consequently, we make the following prediction:
\begin{description}
 \item \textbf{P1}: Participants will rate as warmer a robot initiating body rotation compared to a robot without body rotation.
\end{description}

As stated in \cite{fiske2007universal} warmth relates to the valence of the interaction. Consequently, we expect that participants would prefer a warmer robot, thus, combined with P1, we make the following prediction:
\begin{description}
 \item \textbf{P2}: Participants will prefer a robot initiating body rotation compared to a robot without body rotation.
\end{description}

\subsubsection{Hypotheses on slide}

As show in our observation study and previous work, when crossing each other, people often keep sliding forward \cite{collett1974patterns}. We posit that replicating this human trait would make the robot more organic and social, thus would increase robot's warmth \cite{carpinella2017robotic}. Consequently, we make the following prediction:
\begin{description}
 \item \textbf{P3}: Participants will rate as warmer a robot sliding forward while crossing compared to a robot stopping.
\end{description}

Additionally, as we expect a warmer robot to be preferred, we make the following prediction:
\begin{description}
 \item \textbf{P4}: Participants will prefer a robot sliding forward while crossing compared to a robot stopping.
\end{description}

\subsection{Method}

\subsubsection{Participants}
We recruited and paid 25 participant to take part in the study, 
however two participants were removed due to sensor issues preventing the robot to exhibit a correct behaviour. Consequently, we analyse the results from 23 participants (N=23; age: \textit{M}=29.7, \textit{SD}=12.1; 13 females).

\subsubsection{Environment}

\begin{figure}
    \centering
    \begin{subfigure}[b]{0.30\textwidth}
        \centering
        \includegraphics[width=\textwidth]{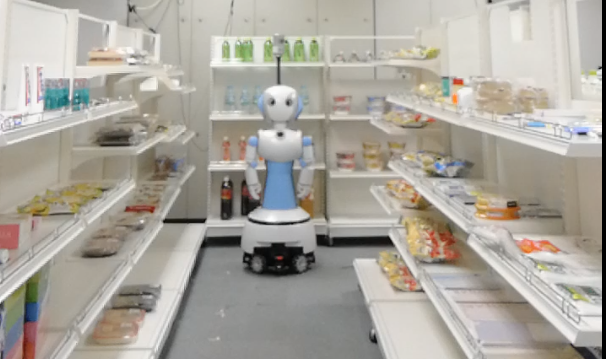}
		\caption{Picture of the robot in one aisle of the shopping environment.}
		\label{fig:photo}
    \end{subfigure}
    \begin{subfigure}[b]{0.17\textwidth}
		\centering
		\includegraphics[width=\textwidth]{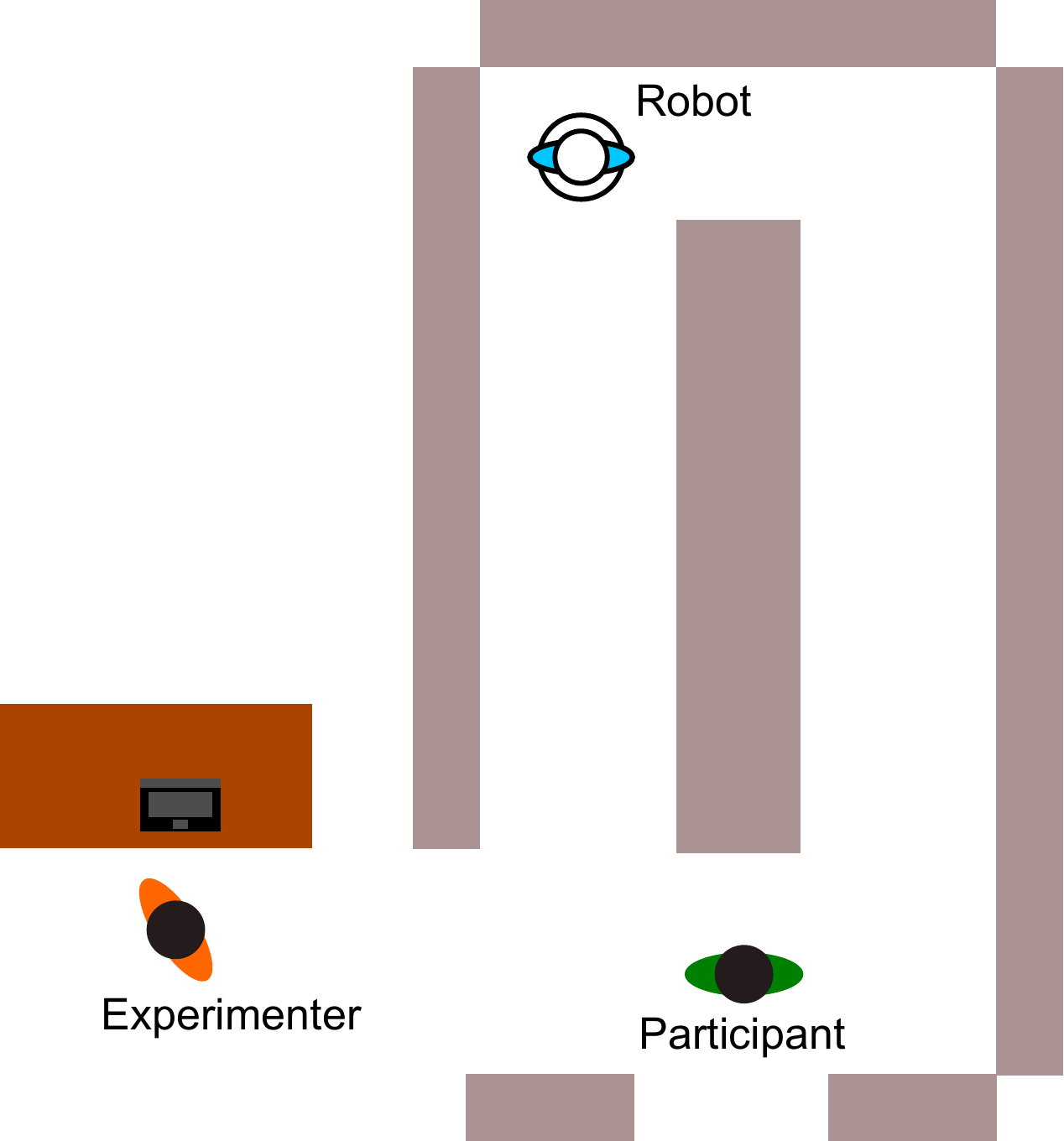}
		\caption{Diagram of the shopping environment.}
		\label{fig:setup-diag}
	\end{subfigure}
    \caption{Setup used in the study.} % The shopping environment consists of two main aisles, where crossing can happen. Before a task, the robot waits for the participant at the top of an aisle.}
    \label{fig:setup}
\end{figure}

This study took place in a laboratory environment simulating a convenient store (cf. Fig. \ref{fig:photo}). 
As shown in Fig. \ref{fig:setup-diag}, the environment consists of shelves surrounding a \emph{circular} empty space people and the robot can navigate in, with additional shelves in the middle.
This leads to two long corridors and two short ones. In our study, all the crossings took place in the long corridors, also named aisles. The aisle's width is 95cm. 

\subsubsection{Conditions}
The study design was 2x2 within participants. Each participants interacted with the robot in all four conditions, and the interaction order was counterbalanced.
Our conditions evaluate the two factors of the \textit{step-slide-rotate} behaviour (implementation is reported in Sec. \ref{sec:system}): the rotation and the sliding.

\paragraph{ \textbf{Rotation factor}} 
The rotation factor has two conditions:
\begin{itemize}
 \item \textbf{Rotation condition}: the robot rotates its body just in time to make space for the incoming participant. 
 \item \textbf{No-rotation condition}: the robot keeps facing forward when crossing with a participant.
\end{itemize}

\paragraph{\textbf{Sliding factor}} 
The sliding factor also has two conditions:
\begin{itemize}
 \item \textbf{Slide condition}: the robot moves towards the participant when crossing with them.
 \item \textbf{No-slide condition}: the robot stops and waits for the participants to pass behind it.
\end{itemize}

In every condition, the robot performed the initial avoiding \textit{step}.

 \begin{figure}
    \centering
	\begin{subfigure}[b]{0.15\textwidth}
	    \centering
		\includegraphics[width=.9\textwidth]{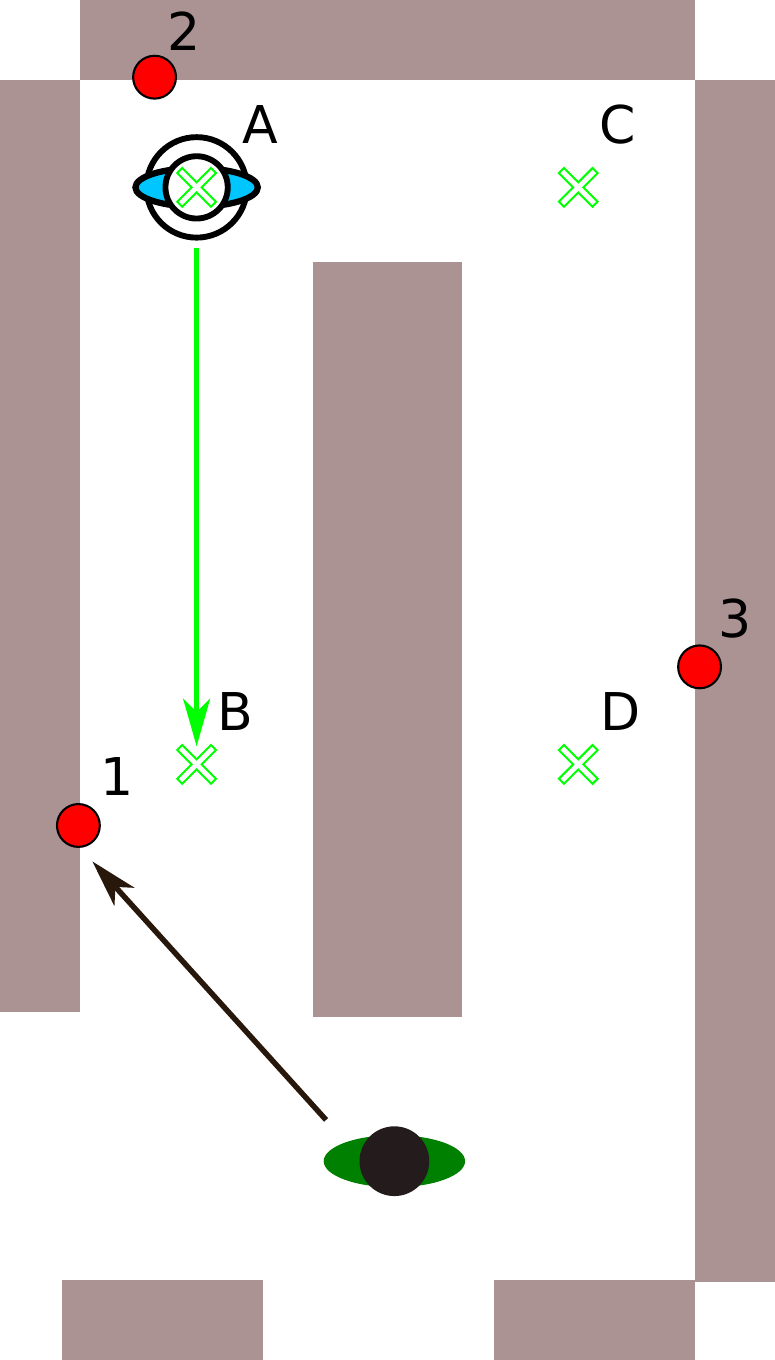}
		\caption{Task 1.}
	\end{subfigure}
	\begin{subfigure}[b]{0.15\textwidth}
	    \centering
		\includegraphics[width=.9\textwidth]{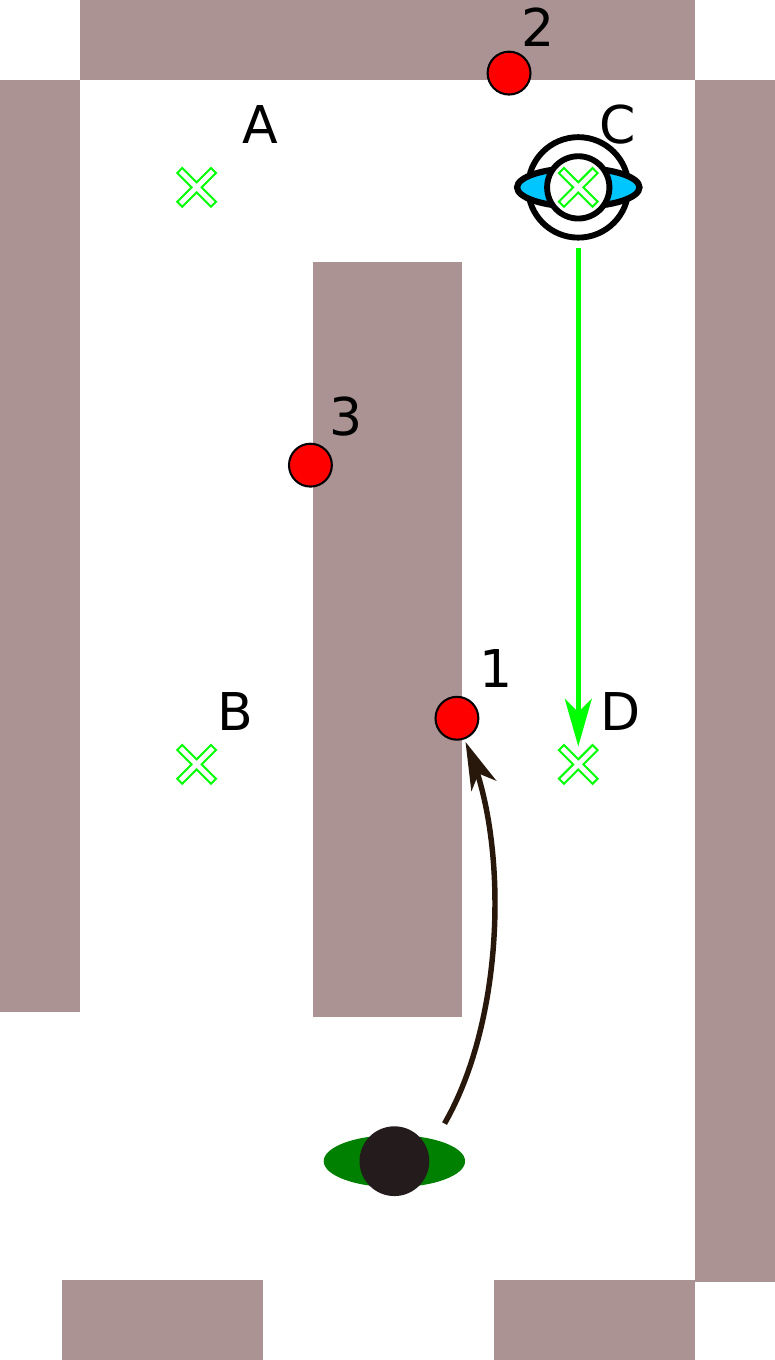}
		\caption{Tasks 2.}
	\end{subfigure}
	\begin{subfigure}[b]{0.15\textwidth}
	    \centering
		\includegraphics[width=.9\textwidth]{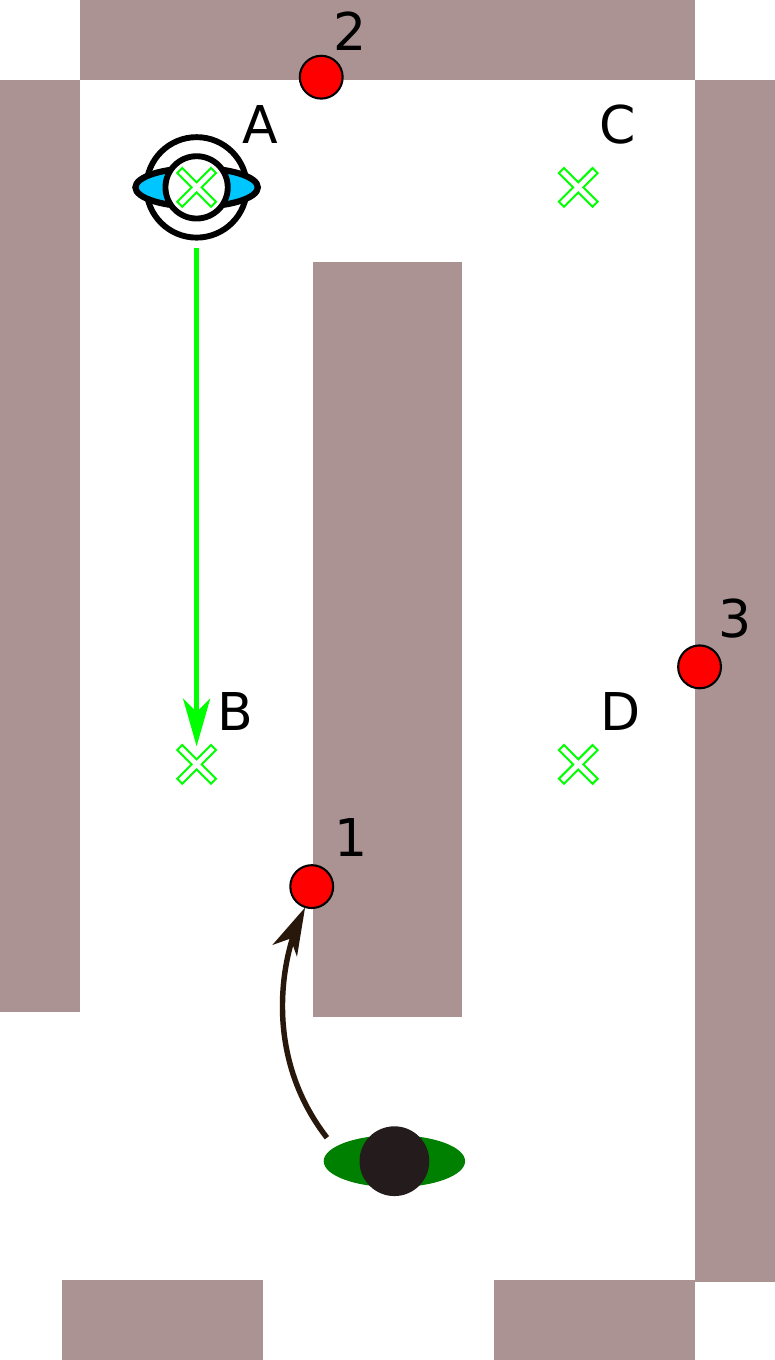}
		\caption{Task 3.}
	\end{subfigure}
    \caption{Diagrams of the three shopping tasks. Participants have to pick up items (located in the red circles) in order, so that they cross the robot on the left aisle in task 1 and 3, and right aisle for task 2. 
    }
    \label{fig:task}
\end{figure}
\begin{figure*}[ht]
	\centering
	\begin{subfigure}[b]{0.18\textwidth}
		\centering
		\includegraphics[width=0.7\textwidth]{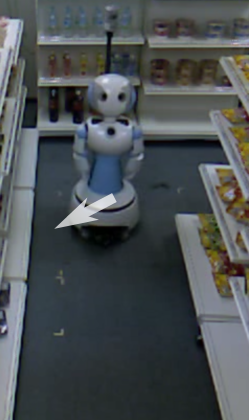}
		\caption{Robot detects the human and steps aside.}
	\end{subfigure}
	\begin{subfigure}[b]{0.18\textwidth}
		\centering
		\includegraphics[width=0.7\textwidth]{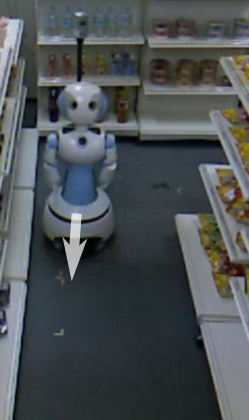}
		\caption{Robot slides forward.\newline}
	\end{subfigure}
	\begin{subfigure}[b]{0.18\textwidth}
		\centering
		\includegraphics[width=0.7\textwidth]{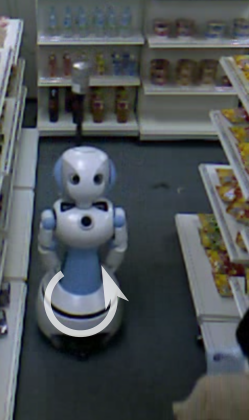}
		\caption{Robot rotates its body.\newline}
	\end{subfigure}
	\begin{subfigure}[b]{0.18\textwidth}
		\centering
		\includegraphics[width=0.7\textwidth]{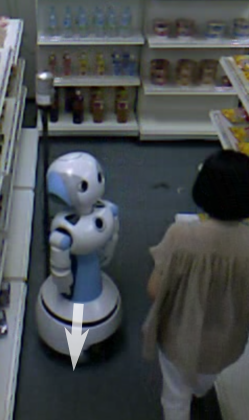}
		\caption{Robot keeps moving forward.}
	\end{subfigure}
	\begin{subfigure}[b]{0.18\textwidth}
		\centering
		\includegraphics[width=0.7\textwidth]{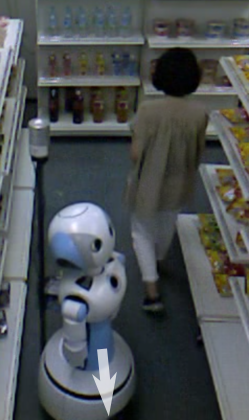}
		\caption{Robot returns to middle.}
	\end{subfigure}
	\caption{Example scenes from the slide and rotation condition. Arrows indicate direction of motions.}
	\label{fig:step_slide}
\end{figure*}

\begin{figure*}[ht]
	\centering
	\begin{subfigure}[b]{0.18\textwidth}
		\centering
		\includegraphics[width=0.7\textwidth]{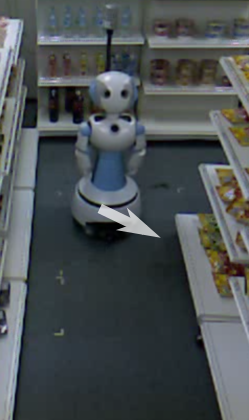}
		\caption{Robot detects the human and steps aside.}
	\end{subfigure}
	\begin{subfigure}[b]{0.18\textwidth}
		\centering
		\includegraphics[width=0.7\textwidth]{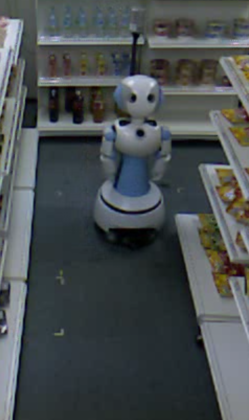}
		\caption{Robot waits.\newline}
	\end{subfigure}
	\begin{subfigure}[b]{0.18\textwidth}
		\centering
		\includegraphics[width=0.7\textwidth]{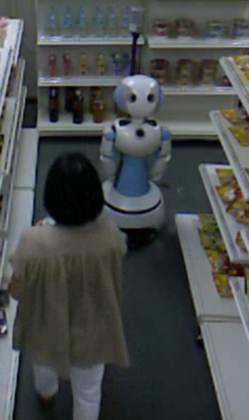}
		\caption{Robot waits.\newline}
	\end{subfigure}
	\begin{subfigure}[b]{0.18\textwidth}
		\centering
		\includegraphics[width=0.7\textwidth]{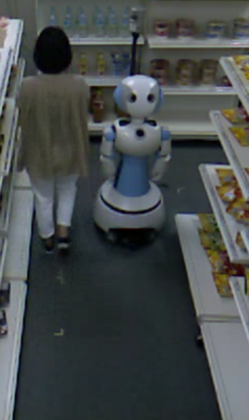}
		\caption{Robot waits.\newline}
	\end{subfigure}
	\begin{subfigure}[b]{0.18\textwidth}
		\centering
		\includegraphics[width=0.7\textwidth]{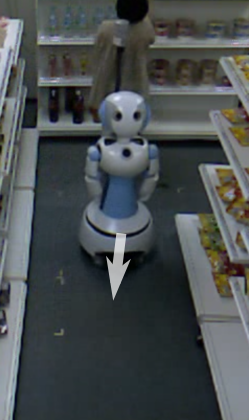}
		\caption{Robot returns to middle.}
	\end{subfigure}
	\caption{Examples scenes from the no-slide and no-rotation condition. Arrows indicate direction of motions.}
	\label{fig:stop_face}
\end{figure*}

\subsubsection{Procedure}
Participants started from an area outside of the simulated shop where they were instructed about their tasks. Participants were not given concrete indication about the robot's objectives, we simply stated that a robot was working in the shop as a shop clerk.

The experiment consisted of four sessions (one per condition) in counter-balanced order. For each session, we set up a shopping situation in which participants had multiple occasions to cross with the robot. That is, each session consisted of three shopping tasks.
For each shopping task, participants had to pick up three specific products in the shop (Fig. \ref{fig:task}).
During the shopping tasks, the robot crossed participants on a long aisle. After each task, the robot moved to the top of the other aisle to be ready for the next one.
After each session, participants completed a questionnaire asking their impressions about the robot, and had a brief interview to report the specificities of this interaction.
After the last session, participants completed a longer interview with the experimenter to compare the difference between the robots in each session.

\subsubsection{Measurement}
After each session, participants filled out a questionnaire based on 1-to-7 point Likert scales. This questionnaire included previously validated scales related to our hypotheses:

\begin{itemize}
    \item \textbf{Warmth} 
    from the RoSAS scale \cite{carpinella2017robotic}, composed of 6 questions (feeling, happy, organic, compassionate, social, and emotional).
    \item \textbf{Likeability} 
    from the Godspeed scale \cite{bartneck2009measurement}, composed of 5 questions (dislike - like, unfriendly - friendly, unkind - kind, unpleasant - pleasant and awful - nice).
\end{itemize}

\section{Results}
\subsection{Observation}
\label{sec:observation}
\begin{figure}
    \includegraphics[width=.15\textwidth]{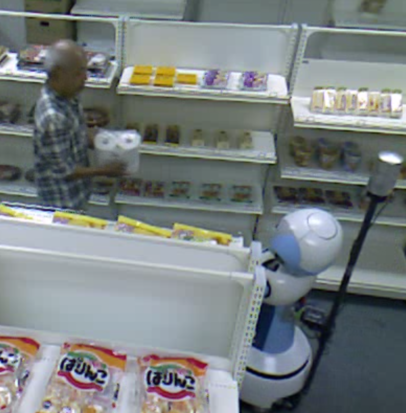}
    \includegraphics[width=.15\textwidth]{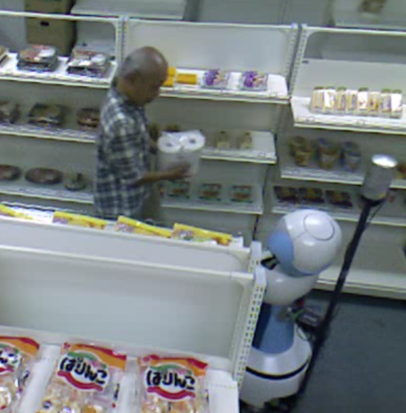}
    \includegraphics[width=.15\textwidth]{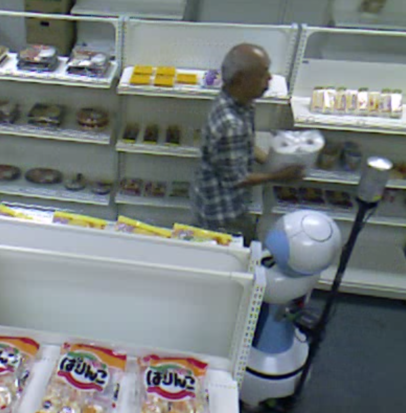}
	\caption{Nodding from one participant to the robot rotating its body.}
	\label{fig:nod}
\end{figure}

\begin{figure}
    \includegraphics[width=.1\textwidth]{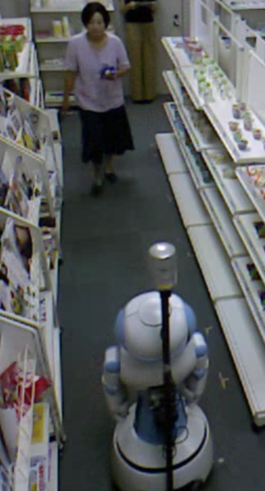}
    \includegraphics[width=.1\textwidth]{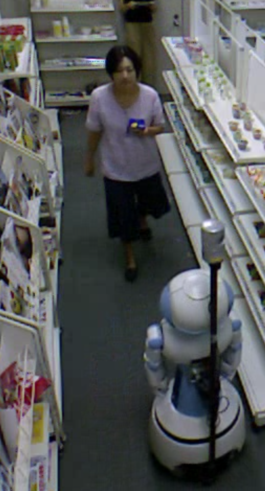}
    \includegraphics[width=.1\textwidth]{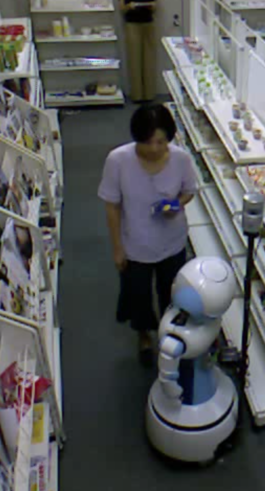}
    \includegraphics[width=.1\textwidth]{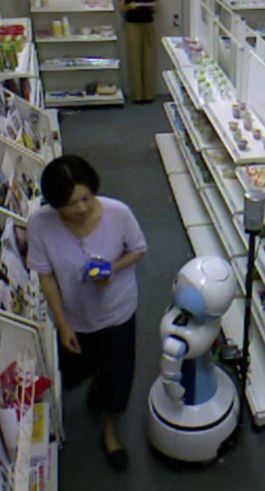}
	\caption{Participant and robot going on the same side.}
	\label{fig:block}
\end{figure}

Fig. \ref{fig:step_slide} shows a typical interaction in the slide and rotation condition: the participant approaches the robot, which side-steps and keeps moving forward until rotating for the crossing (in step c). 
Fig. \ref{fig:stop_face} presents a no-slide and no-rotation condition. In this case, the robot also steps on the side, but does not move anymore until the participant is behind it. 
Participants generally behaved with the robot as people would behave in a shop with other customers, mostly ignoring them. This behaviour tended to be consistent between conditions.
Nevertheless, we did observe some social behaviours towards the robot, some participants smiled or even nodded to the robot (cf. Fig. \ref{fig:nod}). Interestingly, this participant nodded to the robot only when the robot was rotating its body.
We also observed some unusual behaviours from participants. In our study, the robot committed to a side and would keep it. Some participants picked the same side as the robot and tried to make it change side. However, the robot was not designed to do so, and simply stopped in front of the participants, letting them move (cf. Fig. \ref{fig:block}). 

\subsection{Verification of Hypotheses}

\subsubsection{Warmth}

\begin{figure}
		\centering
    \includegraphics[width=0.42\textwidth]{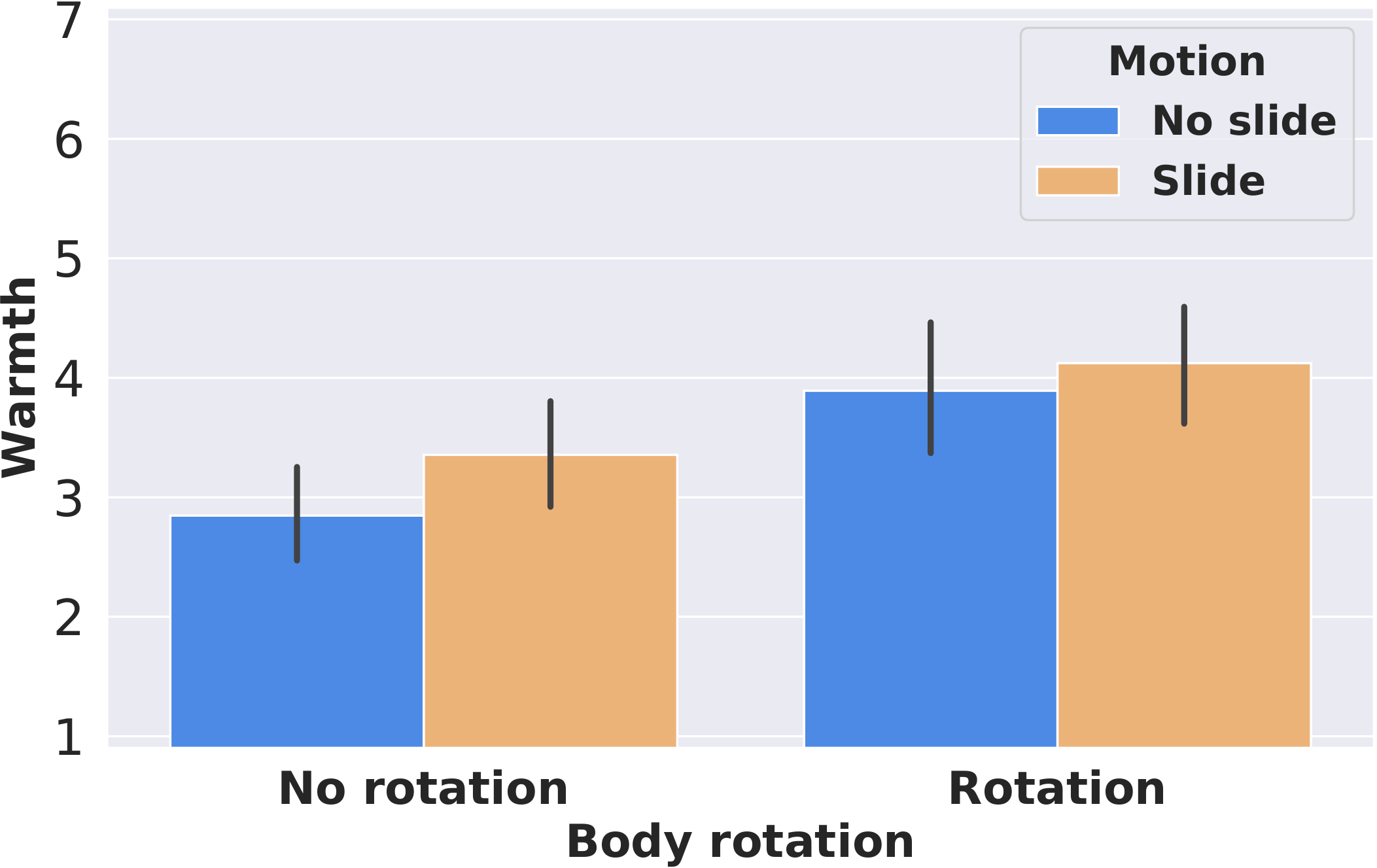}
    \caption{Warmth results from the RoSAS. Bars represent the 95\% confidence interval of the mean (N=23).}
    \label{fig:warmth}
\end{figure}

Fig. \ref{fig:warmth} presents the results from the warmth scale of the RoSAS questionnaire. Repeated-measures ANOVA shows significant effect for the rotation factor ($F(1,22)$ = $32.76$, $p$ $<$ $.001$, %$\eta^2_G$ = $.149$, 
$\eta^2_p$ = $.598$) and the sliding factor ($F(1,22)$ = $4.46$, $p$ = $.046$, %$\eta^2_G$ = $.025$, 
$\eta^2_p$ = $.168$). Their interaction has no significant effect ($F(1,22)$ = $0.74$, $p$ = $.398$, %$\eta^2_G$ = $.003$, 
$\eta^2_p$ = $.033$).
Thus, our predictions P1 and P3 are supported. Participants rated a robot with body rotation warmer compared to a robot without body rotation, and participants rated a robot sliding forward while crossing warmer compared to a robot stopping.

\subsubsection{Likeability}

\begin{figure}
    \centering
    \includegraphics[width=0.42\textwidth]{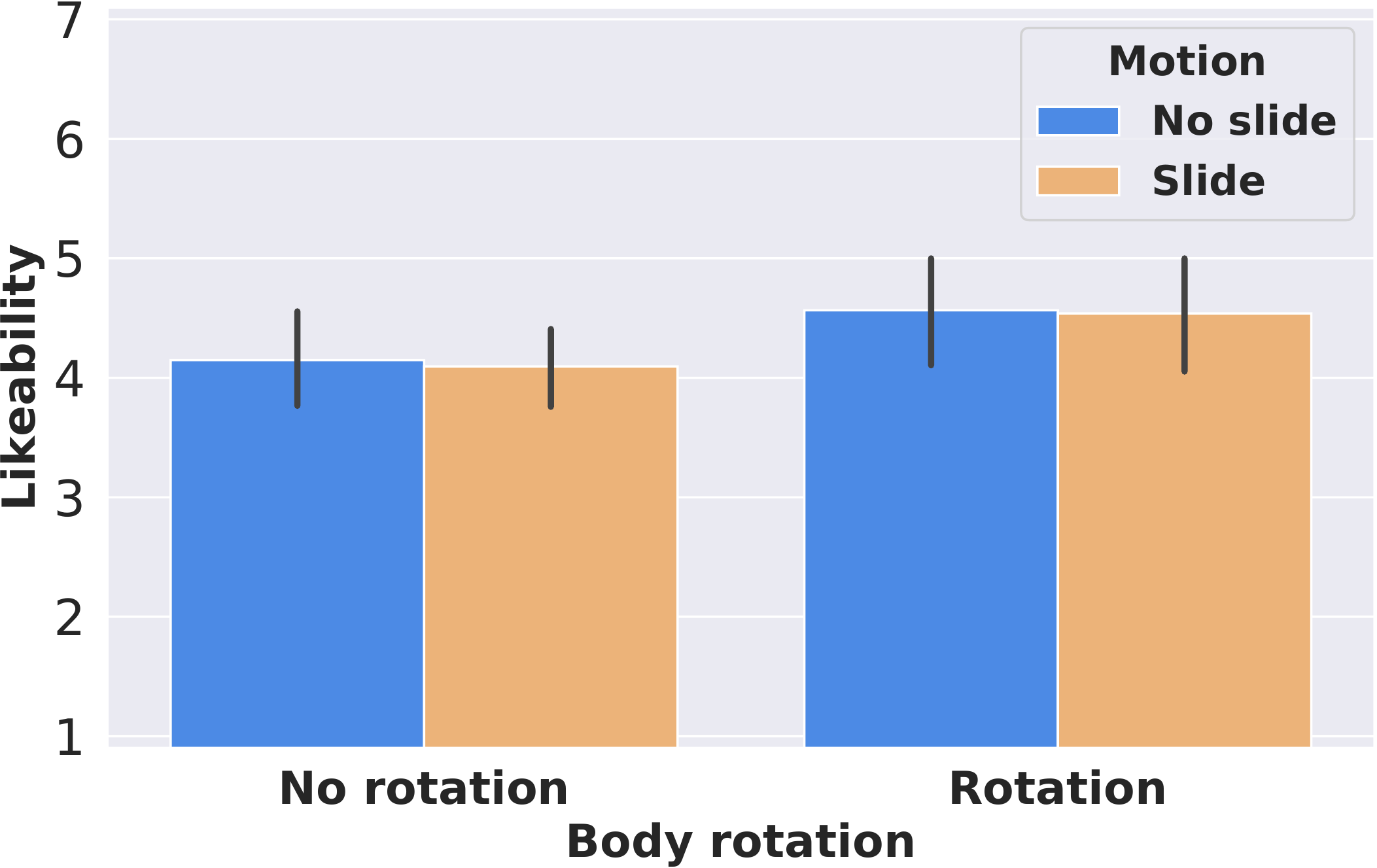}
    \caption{Likeability results from Godspeed. Bars represent the 95\% confidence interval of the mean (N=23).}
    \label{fig:like}
\end{figure}

Fig. \ref{fig:like} presents the results from the likeability scale of the Godspeed questionnaire. Repeated-measures ANOVA revealed the statistically significant effect of the rotation factor ($F(1,22)$ = $7.12$, $p$ = $.014$, %$\eta^2_G$ = $.045$, 
$\eta^2_p$ = $.244$). No significant effect was observed in the sliding factor ($F(1,22)$ = $0.04$, $p$ = $.847$, 
$\eta^2_p$ = $.002$) and their interaction ($F(1,22)$ = $0.01$, $p$ = $.938$, 
$\eta^2_p$ $<$ $.001$).

Thus, our prediction P2 is supported. Participants preferred a robot initiating body rotation compared to a robot without body rotation. On the other hand, our prediction P4 is not supported. 

\subsection{Interviews}

After each interaction with a conditions, participants were asked to give a short description of the robot they just interacted with. Here we report typical descriptions from the participants.

\subsubsection{Slide - Rotate}

In general, participants reported that the slide - rotate robot was looking at them and considering them (e.g.
`'It was kind'', ``It was welcoming the customer'', ``It was caring''). However, for some participants, 
the fact that the robot rotated its body seemed to put some pressure on them, making them feel observed (e.g.
``It wanted to see the customer's face'', ``It was watching human behaviour'').
A limited number of participants did not appreciate the sliding motion, and felt that the robot was not 
helping them (e.g. ``Didn't like the customer'', ``Wanted to disturb'').

\subsubsection{No slide - Rotate}
Similarly to the slide - rotate robot, the no slide - rotate one elicited positive reactions from the participants (e.g.
``It was welcoming the customer'', ``It behaved like a human'', ``Didn't want to disturb the  customer'') but also some pressure due to the gaze and body rotation (e.g. ``It was monitoring'', ``It was worried''). 

\subsubsection{Slide - no rotate}

The participants noticed the sliding behaviour and as the robot was not looking at them, many felt that the robot had an important task to do (e.g. 
``It was rushing'', ``In a hurry'', ``Moving freely''). Participants also reported that the robot was still avoiding them (e.g. 
``It avoided the customer'', ``Wanted to avoid the customer'', ``It tried not to get in the way''), but noted the lack of socialness of the behaviour (e.g.
``It didn't look at me'', ``It was moving mechanically'', ``No interests in human''). Some participants also reported that with the robot, they ``almost bumped into each other'' (the robot would stop when being too close to people, so the risks of actually bumping into humans were very limited).

\subsubsection{No slide - No rotate}
The robot without slide and rotation was characterised by participants by its passivity (e.g.
``It avoided me and stop'', ``It did not look at me'', ``Inorganic''). This passivity had both good effect on participants (e.g.
``It let me pick up the goods'', %``It was caring about the customer'', 
``It was trying not to get in the way''), and bad effects (e.g.
``An obstacle'', ``Wanted to disturb the customer a little''').

\section{Discussion} \label{sec:disucssion}
\subsection{Findings and Interpretation}
The experimental results generally supported our hypothesis. Despite a relatively small sample size, the within-participant design allowed us to find significant results with medium to large effect size. Our {\it step-slide-rotate} model has two important elements, {\it slide} and {\it rotate}. We found that the rotating motion provided warmer impression hence the participants preferred it.
On the other hand, while the slide motion also provided warmer impression, there was no significant difference in preferences. This might be due to the definition of warmth as captured by the RoSAS scale. It includes item such as ``organic'' and ``have emotions'' which are related to the ``alive'' factor of warmth, but not the ``friendly'' factor. In our study, participants interacting with a sliding robot mentioned that the robot looked ``rushed'', while the not-sliding robots could be seen as obstacle and more passive. This difference of agency and potential `threat' felt by some participants interacting with a sliding robot might explain this discrepancy between warmth and likeability.
Despite this absence of preference, if a sliding robot could cover more space in the same amount of time without negatively impacting surrounding people, it could allow it to finish its task faster. This could make a sliding strategy more useful for robot's owners compared to a stopping strategy.

Note that while we imitated people's strategy, one important difference between our robot and people is the shape of the body. People's body is elliptic, so rotation reduces the space they occupy in the orthogonal direction of the corridor. In contrast, our robot has a mostly round shape, so rotating does not change much the space it occupies. We consider it interesting that the orientation of the robot's body still influences people's impression although not affecting their real easiness of passing through. We have not tested other morphologies, but we posit that the effect of the rotation could be more important for robots with more anthropomorphic appearance, such as biped robots.

\subsection{Technical Implications}
There are already commercial robots, e.g. Pepper, that have omni-directional base. Thus, it is possible to use the proposed model into such a robot. However, from our experience in the preparation, there are concerns that need to be carefully addressed when implementing and testing such behaviour. %a couple of issues to be carefully prepared for the implementation.
First, an omni-directional base has a high level of freedom of navigation and control.
In compensation for this freedom, it can easily generate wobbles in navigation due to the interaction with other modules
(such as noise of odometry, laser range finders, resolution of the occupancy map...).
When testing early versions of the software, we noticed that people are highly sensitive to lateral noise motions. For example, variations as small as 5cm could make people feel pressured by the unpredictability of these motions in a narrow space.
Thus, people aiming to use such system should carefully moderate these noises (e.g. applying smoothing filters, reducing the freedom of navigation in problematic cases, reducing velocities, and so on) to have an accurate navigation system comfortable for surrounding humans.

Second, our implementation uses a LiDAR, which provide a high quality mapping of the surrounding space. With the sensors currently used in commercial robots, the quality of sensory inputs could be lower. This could have undesired effects, such as the delay of detection and loss of perception of oncoming person, which could make the robot act in ways making participants uncomfortable.

\subsection{Limitations and Future Works}

While showing the importance of the body rotation for crossing in narrow corridors, this work still possesses limitations.
First, as this research was conducted in Japan, only with Japanese participants, it might be complicated to ensure that similar results would translate in other regions of the world. Especially as it has been shown that proxemics are highly situated in a cultural context \cite{sorokowska2017preferred}. While we acknowledge that the study took place in the lab, and not in the wild, we believe our experiment design captures well the important design factors of narrow corridors navigation and the setup allows to focus solely on the navigation and not other factors such as verbal interaction. Additionally, we believe that the implications of this research go beyond the example used in this study and the specific relation between the robot and human.

Secondly, we only tested ``open'' passes (rotating toward the incoming persons).
However, we found that ``open'' rotation makes the robot's gaze follow the person's position, which provided positive impressions for some of participants (e.g. feeling welcomed), but also induced uncomfortable feelings to some others (e.g. feeling observed).
As people have been shown to practice both open and close rotation \cite{collett1974patterns}, the impact of close rotations should be observed in future work.

Finally, this study aimed to explore the impact of body motion and trajectory solely. However other factors probably have significant effects in the context of navigation in narrow spaces. 
Future work should explore the impact of robot's head gaze, gesture and language on people's perception (e.g. preference, warmth, clarity of intentions, pressure...).

\section{Conclusion}
This paper explored in a human-human pilot study the important factors in human crossing behaviour. Observations showed that both body orientation and sliding motion matter and can vary in such human crossings. We then developed a robot controller allowing to vary these factors and tested it in a within-participant study involving 23 participants. Results from the study showed that both factors impacted the warmth of the robot, and that a robot rotating its body was preferred by participants compared to a non-rotating robot.

To conclude, this paper demonstrated that avoiding humans is not simply about generating a trajectory freeing physical space for people, it is also about following social norms taking into account body orientation.
As such, this paper has two main messages. First, we want to encourage the community to consider both the trajectory and the orientation of robots when navigating in human environments. And second, we think that more research should be made towards rich locomotion tools, such as 
omni-directional locomotion platform, to have more flexibility in robot's navigation control.

\begin{acks}
This work was supported by JST CREST Grant Number JPMJCR17A2, Japan and by the JSPS Postdoctral Fellowships for Research in Japan (Summer Program). Ethical approval was provided by ATR's IRB. We would like to thank the research assistants who helped us to prototype the system, with a special note for Chiaki Kimura who also ran the study.
\end{acks}

\bibliographystyle{ACM-Reference-Format}
\balance
\bibliography{main}

\end{document}